\documentclass[10pt,twocolumn,letterpaper]{article}
\usepackage{cvpr}             
\usepackage{graphicx}
\usepackage{amsmath}
\usepackage{amssymb}
\usepackage{booktabs}
\usepackage{amsfonts}       
\usepackage{nicefrac}       
\usepackage{microtype}      
\usepackage{xcolor}         
\usepackage{multirow}
\usepackage{booktabs}
\usepackage{enumitem}
\usepackage{float}
\usepackage[pagebackref,breaklinks,colorlinks]{hyperref}

\newcommand{\Tref}[1]{Table~\ref{#1}}
\newcommand{\Eref}[1]{Equation~\ref{#1}}
\newcommand{\Fref}[1]{Figure~\ref{#1}}

\usepackage[capitalize]{cleveref}
\crefname{section}{Sec.}{Secs.}
\Crefname{section}{Section}{Sections}
\Crefname{table}{Table}{Tables}
\crefname{table}{Tab.}{Tabs.}
\begin{document}
\title{Self-supervised Transformer for Deepfake Detection} 
\author{Hanqing Zhao\textsuperscript{\rm 1}\qquad  Wenbo Zhou\textsuperscript{\rm 1} \qquad Dongdong Chen\textsuperscript{\rm 2}\\
Weiming Zhang\textsuperscript{\rm 1}\qquad Nenghai Yu\textsuperscript{\rm 1}\\ University of Science and Technology of China\textsuperscript{\rm 1} \qquad Microsoft Cloud AI\textsuperscript{\rm 2}\\
 {\tt\small {\{zhq2015@mail, welbeckz@, zhangwm@, ynh@\}.ustc.edu.cn}} \\ 
{\tt\small {cddlyf@gmail.com}}}

\maketitle

\begin{abstract}
  The fast evolution and widespread of deepfake techniques in real-world scenarios require stronger generalization abilities of face forgery detectors. Some works capture the features that are unrelated to method-specific artifacts, such as clues of blending boundary, accumulated up-sampling, to strengthen the generalization ability. However, the effectiveness of these methods can be easily corrupted by post-processing operations such as compression. Inspired by transfer learning, neural networks pre-trained on other large-scale face-related tasks may provide useful features for deepfake detection. For example, lip movement has been proved to be a kind of robust and good-transferring high-level semantic feature, which can be learned from the lipreading task. However, the existing method pre-trains the lip feature extraction model in a supervised manner, which requires plenty of human resources in data annotation and increases the difficulty of obtaining training data. In this paper, we propose a self-supervised transformer based audio-visual contrastive learning method. The proposed method learns mouth motion representations by encouraging the paired video and audio representations to be close while unpaired ones to be diverse. After pre-training with our method, the model will then be partially fine-tuned for deepfake detection task. Extensive experiments show that our self-supervised method performs comparably or even better than the supervised pre-training counterpart.
\end{abstract}
\section{Introduction}

Face manipulation technologies \cite{suwajanakorn2017synthesizing,FSGAN,kim2018deep,Neuraltexture,koujan2020head2head,chen2020simswap,zhu2021one} empowered by deep generative models are fast advancing which makes deepfake medias more realistic and easily to deceive watchers. The malicious usage and spread of deepfake have raised serious societal concerns and posed an increasing threat to our trust in online media. Therefore, deepfake detection in real-world scenarios becomes an urgent need and obtains a considerable amount of attention in recent years.

To defend against the potential risks of these forged media, numerous efforts have been devoted and achieving promising performances on specific datasets in recent years \cite{ff++,wu2020sstnet,masi2020two,qian2020thinking,faceXray,agarwal2020detecting,madd,liu2021spatial,Lipforensics}. Many previous works introduce low-level texture features from different domains for searching the underlying generation artifacts. However, dramatic drops in performance may be experienced when the artifact patterns are changed. And the well-processed forged videos which show only subtle differences from real ones make the deepfake detection in real-world scenarios to become a very tough task.

Some works attempt to strengthen the generalization ability by data augmentation or capturing common clues during the forgery process. For example, SPSL\cite{liu2021spatial} focuses on finding the frequency artifacts caused by the accumulative up-sampling operation. Another effective way is to predict the blending boundaries \cite{faceXray,zhao2020learning} between the background and the altered inner face regions. Although they achieve impressive performances in cross-data evaluations, they are usually sensitive to post-processing, e.g., video compression. Recently, researchers try to develop robust high-level semantic features. Heliasos et al. \cite{Lipforensics} find that supervised pre-trained spatio-temporal networks with visual speech recognition(lipreading) tasks can extract robust representations of lip movements for boosting deepfake detection performance. This also indicates that feature extractors pre-trained in other face-related tasks can provide meaningful help for deepfake detection. However, pre-training in other tasks such as lipreading requires precise segmentation and data annotation. It is costly when developing a larger training set.

To address the limitation, we propose a self-supervised transformer for pre-training a robust semantic feature extractor. Generally speaking, there is a strong correlation between audio and lip movement in speech videos. It has also been proved that predicting lip movement by audio (lip synthesis \cite{suwajanakorn2017synthesizing}) or predicting audio by lip movement (lip2wav \cite{prajwal2020learning}) are feasible. Therefore, we propose the method to self-supervised pre-train instead of supervised pre-train in the lipreading task. We can obtain a general representation of lip movement by audio-visual consistency using contrastive learning.

To this end, we propose a two-stage spatio-temporal video encoder. In the frontend stage, we use a 3D convolution layer to extract flows from video, followed by a 2D CNN for capturing the local lip movement representations and a temporal 1D transformer backend for long-term lip movement representations. To learn the generic lip movements representation, we design a cross-modal contrastive learning method that utilizes the consistency between the audio channel and the movement of lip regions in frames to train the video encoder. After pre-training, we freeze the front end of the encoder and add an MLP head to fine-tune the whole network for deepfake detection task.

We conduct extensive experiments to compare the performances of our methods with the state-of-the-art methods in various challenging cases. The results demonstrate that our method achieves comparable performance with the state-of-the-art in most cases, and achieves better performance with respect to generalization to unseen forgery datasets. Compared to the supervised pre-trained method, our approach also exhibits better robustness to common corruptions which degrade other models' performance. Further, we investigate the relation between the scale of pre-training data and the average detection AUC. Conclusively, a larger pre-training data scale is indeed helpful to promote the performance of models. Another advantage of our method is significantly reducing the annotation cost for pre-training data.

\section{Related Work}

\subsection{Deepfake Detection}

Since the deepfake technique has caused severe societal concerns, many effective countermeasures \cite{masi2020two,qian2020thinking,faceXray,agarwal2020detecting,madd,liu2021spatial,Lipforensics,wang2021m2tr,chen2021local} have been proposed against it. According to the feature extraction types, current methods can be roughly categorized into two types: textural feature based methods and semantic feature based methods. As the name implies, textural feature based methods focus on capturing low-level textural information from different domains. Two-branch\cite{masi2020two} introduced an extra CNN stem with deep Laplacian-of-Gaussian filter in the CNN-RNN architecture for acting as a band-pass filter to amplify artifacts. Patch-forensics\cite{chai2020makes} proposed a classifier with limited receptive fields that focus on textures in small patches to captured local errors. Multi-attention \cite{madd} attempt to introduce the multi-regional attention mechanism into deepfake detection, which is inspired by fine-grained classification. 

To boost the generalization ability, Face X-ray \cite{faceXray} and Patch-wise Consistency Learning \cite{zhao2020learning} leverage the clues of image blending between the background and the altered face region. These methods achieve impressive performances in cross-data evaluation. However, they are easily influenced by video compression and noise disturbance since the boundary clues are highly fragile to post-processing.

Recently, researchers find that high-level semantic features show excellent robustness in dealing with both cross-data tests and post-processing operations. Lipforensics \cite{Lipforensics} extract the representations of lip movement by using a pre-trained network. The lip feature extractor is pre-trained in a supervised manner on Lipreading in the Wild(LRW)\cite{Chung16} dataset, which is commonly used for the lipreading tasks. The network takes a multi-branch temporal convolution network(MSTCN)\cite{mstcn} as backend and fine-tuned on deepfake datasets. It achieves state-of-the-art generalization ability. However, to pre-train such a robust lip feature extractor requires a large-scale well-annotated dataset, which is extremely costly. In this paper, we propose a self-supervised transformer for pre-training, in this way, we can significantly reduce the annotation cost for pre-training data and obtain good scalability.

\subsection{Contrastive Learning}

Recently, general pre-trained models have been widely used for fine-tuning on downstream tasks. Self-supervised pre-training attracts tremendous attention for its versatility. Self-supervised learning enables us to learn meaningful representation for various classification tasks without relying on labeled data. Contrastive learning is a popular self-supervised strategy that encourages the features of the same instances to be close while features of different instances to be distant. Generally, we can take different views of the same sample data as positive pairs and views from unmatched data as negative pairs to construct a classification task for discriminating instances. For pre-training semantical image representations, \cite{Chen2020ASF,Chen2021AnES} obtain different views of the same image by random augmentations. For pre-training video classification models\cite{Feichtenhofer2021ALS}, positive pairs are fragments from the same video while negative pairs are fragments from different videos. Besides the uni-modal contrastive learning, natural multi-media data have cross-modal consistency. 

Many previous self-supervised learning methods are designed in the cross-modal predictive way, while it is also suitable for using contrastive learning that takes each modal as different views with shared information. For example, Ommer et al.\cite{Sayed2018CrossAL} proposed contrastive learning for the frame and optical flow in videos. \cite{CLIP} demonstrate a simple pre-training task, that is, predicting which caption goes with which image is an efficient and scalable way to learn good image representations. Some recent works\cite{Zolfaghari2021CrossCLRCC,avid,Ma2021cont} leverage correspondence between audio and visual signals for learning semantic representations, they achieve strong transfer learning performance on downstream tasks like action recognition and lipreading. Inspired by previous successful works, in this paper, we try to learn a general representation of lip movement by audio-visual consistency using contrastive learning.

\begin{figure*}
   \centering
   \includegraphics[width=\linewidth]{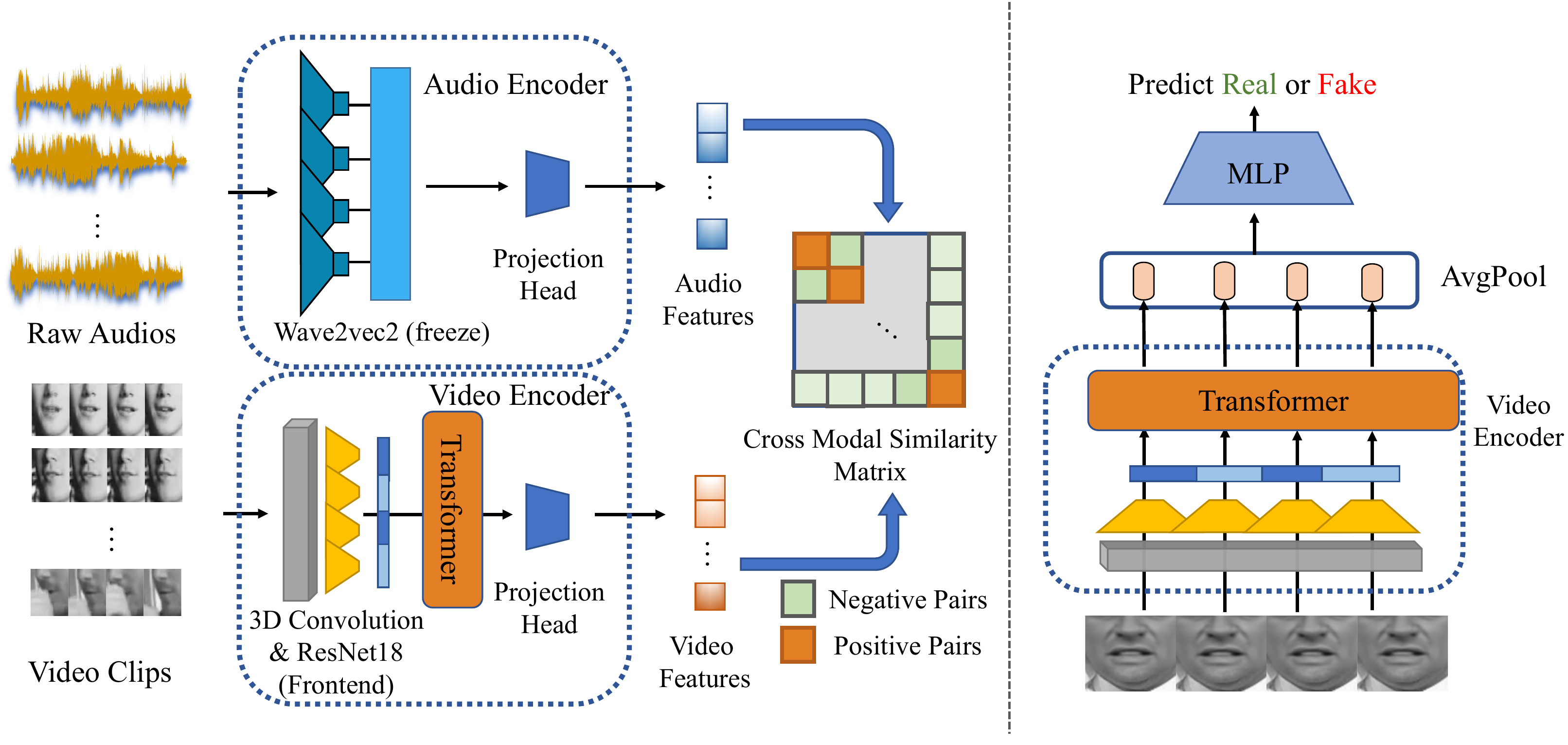}
   \caption{Demonstration of pre-training stage and fine-tuning stage of proposed method. The left part shows the procedure of audio-visual contrastive learning for obtaining generic representation of lip movements. And the right part shows how do we fine-tune the pre-trained video encoder for deepfake detection task. }
   \label{fig_arch}
 \end{figure*}
 
\section{Proposed Method}
\subsection{Overview}

Most deepfake generation methods synthesize fake faces frame by frame without considering motion coherence, especially in the lip regions where movements are most frequent and complex while talking. Thus the irregularity of lip movement can be a common feature of deepfake video and such a high-level semantic feature is robust to various video post-processing. To learn semantical representations of lip movements, we apply cross-modal contrastive learning by encoding utterance features and lip motion features from videos into a common space. The we use instance discrimination learning to enforce corresponding audio and visual features matching each other, which shares a similar spirit with the lipreading task. The learned lip representations can transfer to expose the irregularities in deepfake videos. Benefit from the primitive features of lip movements, the video encoder would be less prone to overfitting non-transferable artifacts when fine-tuning on deepfake detection task.

\subsection{Audio-Visual Contrastive Learning pre-train}

For cross-modal contrastive learning, our model consists of a video encoder and an audio encoder. In the pre-train stage, each encoder uses an additional MLP head for projecting the raw features into the common space. 
We take a pre-trained wav2vec2 model\cite{Baevski2020wav2vec2A} for speech recognition as the backbone of the audio encoder. The wave2vec2 model encodes raw audio into a sequence of local wave features by one dimensional CNN layers which are frozen during training and then inference the features by transformer layers to get semantical audio representation array. We apply temporal adaptive average pooling for converting the output feature array from the last transformer layer to a fixed length token sequence, then the sequence is reshaped as the input of the MLP projection head.

We design a spatial-temporal framework as shown in figure \Fref{fig_arch} for encoding video features. It has two stages, in the frontend stage, a 3D convolutional layer with kernel size [5,7,7] extracts the optical flows from raw videos, then a following 2D ResNet module encodes the local lip movement feature from the optical flows. In the backend stage, we convert the local lip movement feature of each frame into a feature sequence with linear projection, then we use a temporal transformer to get the global semantic video representation. Just like the audio encoder, we also apply temporal adaptive average pooling and MLP projection head for encoding the final video feature.

Previous contrastive learning methods for video representations \cite{avid,Ma2021cont,tclr,videomoco} usually leverage 3D CNN architectures for general video understanding. Compared to 3D CNN, our architecture separates the process of encoding local lip movement feature and global semantic video representation. Our framework makes the fine-tuning process easier because we can freeze the frontend for keeping the low-level feature representations learned during the pre-training, only fine-tune the backend for capturing the long-term inconsistency of fake videos. 

In our framework, we use the transformer as the backend of the encoder. Transformer has succeeded in many NLP and vision tasks since it can capture long-term dependency and preserve less inductive bias than CNN. Thus we believe it is more suitable for large-scale pre-trained models. The shape of output tensor from frontend is noted as $T\in{B\times L \times N}$, $B$ is the batch size, $L$ is a static sequence length with padding, $N$ is the dimension of pooled spatial feature, for example, $N$ is 512 for ResNet18 frontend. We use a linear projection for transforming $T$ to the input features of the transformer, and we add learnable positional embeddings to input features for introducing time relationships. In the experiment part, we validate the model with transformers backend can achieve better performance than those with general CNNs.

We utilize the intrinsic correlation of lip movement and speech audio to learn generic representations of them. Following the contrastive learning manner, we use the InfoNCE\cite{infoNCE} loss to distinguish positive audio-visual feature pairs from a bunch of negative pairs. Video encoder $E_v$ encodes a video clip into video embedding $m_v$ and similarly audio encoder $E_a$ encodes an audio clip into audio embedding $m_a$. Then $m_v$ and $m_a$ are projected to common space by MLP heads as $z_v$ and $z_a$. The loss is shown in equation \Eref{equ:infonce}, we respectively take the video features and the audio features as query and the other modal as keys to get loss $\mathcal{L}_{va}$ and $\mathcal{L}_{av}$, then we simply use their average as the contrastive loss.

\begin{equation}
\mathcal{L}_{va}=-\log \frac{\exp \left(z_v \cdot {z_a^+} / \tau\right)}{\exp \left(z_v \cdot {z_a^+} / \tau\right)+\sum_{{z_a^-}} \exp \left(z_v \cdot {z_a^-} / \tau\right)}
\end{equation}
\begin{equation}
\mathcal{L}_{av}=-\log \frac{\exp \left(z_a \cdot {z_v^+} / \tau\right)}{\exp \left(z_a \cdot {z_v^+} / \tau\right)+\sum_{z_v^-} \exp \left(z_a \cdot {z_v^-} / \tau\right)}
\end{equation}
\begin{equation}
\mathcal{L}=\frac{1}{2}\mathcal{L}_{av}+\frac{1}{2}\mathcal{L}_{va}
\label{equ:infonce}
\end{equation}

Here the label $^+$ means positive sample pairs and $^-$ means negative sample pairs, $\tau$ is the temperature which is set 0.1. For each iteration in the pre-train stage, we sample a mini-batch of video clips, extract segments of synchronized frames and waves with random offset for each video clip as positive audio-visual pairs while all the unrelated items from the mini-batch act as negative pairs.

\subsection{Fine-tuning on Deepfake Dataset}

In the fine-tuning stage, we freeze the parameters in the frontend 3D convolution and ResNet2D for keeping the pre-trained local representations. We extract global features by adaptive average pooling the output of the last transformer layer and adding a new MLP head for classification. 
To preserve the pre-trained knowledge in transformer layers, we manually control the learning rate of each transformer layer. The MLP module is assigned with an initial learning rate of 0.01,  while the last transformer layer is assigned with an initial learning rate of 0.005, each prior transformer layer decays the initial learning rate by 0.9. 
For deepfake detection task, we use binary cross-entropy loss to fine-tune the classifier.
\begin{equation}
   \mathcal{L}_{ce}=y\log(\frac{1}{1+e^{-C(x)}})+(1-y)\log(1-\frac{1}{1+e^{-C(x)}})
\end{equation}
where $x$ and $y$ are the input video sequence and corresponding label, respectively. $C(x)$ represents the predicted logit of the MLP module.

To mitigate the influence of different frame rates, we convert all videos to 25 fps and all audios to a 16kHz sample rate. Then we trace the face of the speaking person with RetinaFace \cite{Deng2019RetinaFaceSD} and SyncNet \cite{Chung2016OutOT}. We crop the selected faces with a tight boundary and use FAN \cite{fan} to get the 68 landmarks. We align each face using 5-landmarks and resize the face images to 256x256. We then crop a 96x96 region at the center of the lip landmarks region. After preprocessing, a video's lip regions can be represented as $F_v \in R^{[T, H, W]}$, where $T$ is the length in time and $H, W$ are height and width. In the pre-train stage, we randomly sample a clip with fixed length $L_p$ and corresponding raw audio with length $640L_p$ from the video. We apply identical image augmentations on each frame including random crop to 88x88, motion blur, Gaussian noise and random brightness. 

The computation complexity of self-attention is the square of the sequence length that forbids us to predict with a very long video sequence. For efficiency, we chunk videos into clips with a smaller length $L_f$ and use the same augmentations as it in the pre-train stage for fine-tunning. We use 2-second clips of videos for pre-training and only 1-second clips of videos for deepfake detection, the longer sequence in contrastive learning can diminish inner-modal similarity to reduce impacts of false-negative pairs and a shorter sequence can speed up the inference procedure. It is worth mentioning that the local lip movement feature extracted by the frontend is irrelevant to the sequence length.

\section{Experiments}

\begin{table*}[t]
\begin{center}
\setlength{\tabcolsep}{5.3mm}{
\begin{tabular}{l c c c c c c c}\hline
\multirow{2}{*}{Method} & \multirow{2}{*}{pre-train} & \multicolumn{3}{c}{Video-level ACC (\%)} & \multicolumn{3}{c}{Video-level AUC (\%)} \\ \cmidrule(lr){3-5} \cmidrule(lr){6-8}
 & & Raw & HQ & LQ & Raw & HQ & LQ \\ \hline 
Xception \cite{ff++} & - & 99.0 & 97.0 & 89.0 & 99.8 & 99.3 & 92.0 \\
CNN-aug \cite{wang2020cnn} & - & 98.7 & 96.9 & 81.9 & 99.8 & 99.1 & 86.9  \\
Patch-based \cite{chai2020makes} & - & 99.3 & 92.6 & 79.1 & 99.9 & 97.2 & 78.3 \\
Two-branch \cite{masi2020two} & - & --- & --- & --- & --- & 99.1 & 91.1 \\
Face X-ray \cite{faceXray} & Sup-BI & 99.1 & 78.4 & 34.2 & 99.8 & 97.8 & 77.3 \\
CNN-GRU \cite{sabir2019recurrent} & - & 98.6 & 97.0 & 90.1 & 99.9 & 99.3 & 92.2 \\
LipForensics \cite{Lipforensics} & Sup-LRW & 98.9 & \textbf{98.8} & \textbf{94.2} & 99.9 & \textbf{99.7} & \textbf{98.1} \\ 
Lipforensics* \cite{Lipforensics} & Sup-LRW & 98.6 & 96.7 & 88.6 & 99.8  & 99.3 & 94.9  \\ \hline
ours & Self-Sup & \textbf{99.2} & 98.6 & 93.5 & \textbf{99.9} & 99.6 & 96.7 \\
\hline
\end{tabular}
}
\end{center}
\vspace{-1em}
\caption{Performance when fine-tune the model on each compression rate of FF++ dataset and test for the same compression rate. Our model achieves competitive performances on raw videos and HQ videos.}
\label{table:ff++}
\end{table*}

\subsection{Implement Details}

We use $VoxCeleb2$ dataset \cite{voxceleb2} and part of $AVSpeech$ dataset \cite{Ephrat2018LookingTL} for pre-training, totally contains 2,800,000 speech video clips. Each video clip includes more than 100 continuous frames with a trackable face. We use AdamW optimizer for pre-train, the initial learning rate is 0.01 and decay 0.9 for each epoch, the minimum learning rate is 0.0001. The hyper-parameter of each transformer layer is 1024 dims, 8 attention heads, 128 dims for each attention head, 2048 intermediate dims and 0.2 dropout rate. The dimension of features for contrastive learning is 256. We end the pre-training when the training loss stops descending for 3 epochs. We conduct extensive experiments with different hyper-parameter combinations to determine the optimal selection for our method. Parts of the results are listed in \Tref{table:ab_layers}. Finally, we choose ResNet18 as frontend and 6 layers transformer as backend for the balance of efficiency and accuracy.

We fine-tune our pre-trained video encoder on original FaceForensics++ dataset\cite{ff++}  (FF++, 4 generation methods: Deepfake, FaceSwap, Neuraltextures, Face2Face and Original videos, 720 videos for each class).  And we use the test set of Celeb-DF v2 \cite{Li2020CelebDFAL} (518 videos), Deeper-forensics \cite{Jiang2020DeeperForensics10AL} (140 videos, paired with original videos of FF++ ), 140 Faceshifter videos from FF++ (paired with original videos)  and DFDC \cite{Dolhansky2020TheDD}  (3000 videos selected from the test set, exclude extremely corrupted videos) for evaluating the cross dataset ability. Following the commonly used evaluation metrics in previous deepfake detection works, we leverage video level ROC-AUC score and binary classification accuracy for the evaluation of the detection performances. 

Notably, due to the training set being unbalanced (the fake videos are 4 times as many as real videos), previous methods usually over-sample real videos. In this work, we sample real videos with similar quality from the pre-train dataset to make the training set balanced.

\subsection{Evaluation on FaceForensics++}

FaceForensics++ is the most widely used dataset for deepfake detection. The original videos of the FF++ dataset include speeches by people of different races. The dataset chooses similar identity pairs for face-swapping, a total of five face swapping methods are used to generate face forgeries on each pair. To consist with previous works, we only use 4 face forgery methods: Deepfake, FaceSwap, Neuraltextures, Face2Face to fine-tune our models, and we treat the FaceShifter as a discrete dataset. There are three compression rates of the FF++ dataset, uncompressed (Raw), slightly compressed (HQ), and heavily compressed (LQ). The uncompressed dataset is relatively easy for spotting deepfake artifacts, but the challenge is increasing with compression rate, in the LQ videos, most textural features are lost which is extremely challenging for detection.

In \Tref{table:ff++}, we report the video-level AUC and accuracy for each compression rate compared to other methods. Among which, Xception \cite{ff++}, CNN-aug \cite{wang2020cnn}, Patch-based \cite{chai2020makes}, Two-branch \cite{masi2020two} and CNN-GRU \cite{sabir2019recurrent} are networks without special pre-train strategies. Face X-ray \cite{faceXray} generates training data by blending real faces and being supervised with blending boundary regression. Lipforensics \cite{Lipforensics} is pre-trained in a supervised manner on Lipreading in the Wild (LRW)\cite{Chung16} dataset. To parallel compare our self-supervised method to the supervised one, we also use the same network structure and pre-train it on LRW with the same training strategy of Lipforensics in a supervised manner, the model is denoted as Lipforensics*. Apparently, our self-supervised pre-trained model outperforms the supervised one.

From the results in \Tref{table:ff++}, we can observe that our model pre-trained in self-supervised manner performs much better than it in supervised manner. Compared to the state-of-the-art Lipforensics method, our model also achieves comparable performances on raw videos and HQ videos while slightly dropping on LQ videos.

\subsection{Evaluation of Cross-manipulation Ability}

Generalizing to unseen forgery classes is challenging in deepfake detection because the artifact patterns are different in each manipulation methods. Some deepfake detection methods trained with several forgery classes may overfit multiple class-specific artifacts but not capture the common feature transferable to new method. 

In this experiment, we evaluate the models' generalizaiton ability cross manipulation methods with the leave-one-out strategy. Each time, we train the model with 3 forgery classes in FF++ HQ dataset and test with the remaining forgery class. The results are shown in  \Tref{table:manip_general}. Our method achieves better performance than most baseline methods, this indicates the lip movement features captured by our model are well generalizable.

\begin{table}[t]
\begin{center}
\resizebox{1\linewidth}{!}{
\begin{tabular}{l c c c c c}\hline
\multirow{2}{*}{Method} & \multicolumn{4}{c}{Train on remaining three} \\  
\cmidrule(lr){2-5}
& DF & FS & F2F & NT & \textbf{Avg} \\ \hline
Xception \cite{ff++} & 93.9 & 51.2 & 86.8 & 79.7 & 77.9 \\
CNN-aug \cite{wang2020cnn} & 87.5 & 56.3 & 80.1 & 67.8 & 72.9 \\
Patch-based \cite{chai2020makes} & 94.0 & 60.5 & 87.3 & 84.8 & 81.7 \\
Face X-ray \cite{faceXray} & 99.5 & 93.2 & 94.5 & 92.5 & 94.9 \\
CNN-GRU \cite{sabir2019recurrent} & 97.6 & 47.6 & 85.8 & 86.6 & 79.4 \\
LipForensics \cite{Lipforensics} & \textbf{99.7} & 90.1 & \textbf{99.7} & \textbf{99.1} & \textbf{97.1} \\
LipForensics* & 97.8 & 90.5 & 98.0 & 96.9 & 95.8 \\ \hline
ours & 98.5 & \textbf{91.9} & 98.3 & 96.4 & 96.3 \\ \hline
\end{tabular}
}
\end{center}
\vspace{-1em}
\caption{\textbf{Cross manipulation method generalisation.} Video-level AUC (\%) when testing on each forgery type of FaceForensics++ HQ after training on the remaining three. The types are Deepfakes (DF), FaceSwap (FS), Face2Face (F2F), and NeuralTextures (NT).}
\label{table:manip_general}
\end{table}

\subsection{Evaluaton of Cross-datasets Generalization}

In real deepfake detection scenarios, the distribution of deepfakes is more complex. The forged videos are not only diverse in source videos and generation methods, there is also a diversity in post-processing methods. Thus the domain generalization ability is a significant metric for deepfake detection models. Since there are large domain gaps between different deepfake datasets, in this part, we evaluate the domain generalization ability of deepfake detectors by cross-dataset test. We fine-tune our model on the FF++ training set with four manipulation methods (DF, F2F, FS, NT), and report the AUC scores tested on Celeb-DF, DFDC, FaceShifter and DeeperForensics dataset, respectively.

In \Tref{table:cross_dataset}, our method achieves state-of-the-art performances in generalization to Celeb-DF-v2, DFDC and FF++ FaceShifter dataset. Our self-supervised pre-trained model surpasses Lipforensics and the same model using supervised training strategy ( Lipforensics*) for an average AUC of 0.7 percent. The performances on FaceShiter and Deeperforensics are obviously better than those on Celeb-DF-v2 and DFDC. It is probably because these two datasets share the same original videos with the FF++.

\subsection{Ablation Study}

\begin{table}[t]
\begin{center}
\resizebox{1\linewidth}{!}{
\begin{tabular}{l c c c c c}\hline
Method & CDF & DFDC & FSh & DFo & \textbf{Avg}  \\ \hline
Xception \cite{ff++} & 73.7 & 70.9 & 72.0 & 84.5 & 75.3  \\
CNN-aug \cite{wang2020cnn} & 75.6 & 72.1 & 65.7 & 74.4 & 72.0 \\
Patch-based \cite{chai2020makes} & 69.6 & 65.6 & 57.8 & 81.8 & 68.7 \\
Face X-ray \cite{faceXray} & 79.5 & 65.5 & 92.8 & 86.8 & 81.2 \\ 
CNN-GRU \cite{sabir2019recurrent} & 69.8 & 68.9 & 80.8 & 74.1 & 73.4 \\
Multi-task \cite{nguyen2019multi} & 75.7 & 68.1 & 66.0 & 77.7 & 71.9 \\
DSP-FWA \cite{li2019exposing} & 69.5 & 67.3 & 65.5 & 50.2 & 63.1 \\
LipForensics \cite{Lipforensics} & 82.4 & 73.5 & 97.1 & \textbf{97.6} & 87.7 \\ 
LipForensics* & 83.5 & 73.7 & 96.3 & 97.2 & 87.7 \\ \hline
ours & \textbf{84.2} & \textbf{74.5} & \textbf{97.8} & 97.3 & \textbf{88.4} \\ \hline 
\end{tabular}
}
\end{center}
\vspace{-1em}
\caption{\textbf{Cross-dataset generalisation.} Video-level AUC (\%) on Celeb-DF-v2 (CDF), DeepFake Detection Challenge (DFDC), FaceShifter HQ (FSh), and DeeperForensics (DFo) when trained on FaceForensics++.}
\label{table:cross_dataset}
\end{table}

\subsubsection{Determination of Model Architecture}

In Lipforensics, the network architecture is referred from a lipreading model\cite{ma2020towards}. In this paper, we propose a self-supervised audio-visual contrastive pre-train instead of a lipreading pre-train. Different from lipreading tasks that classify limited words, audio-visual contrastive learning requires the model to extract more detailed semantic features. Thus a transformer would be a better choice than MSTCN in lipreading tasks. The results in \Tref{table:ab_layers} also demonstrate that transformer performs better than MSTCN with the same layer number. We also conduct a group of experiments to explore a relatively better parameter scale of the frontend 2D ResNet and the backend transformer. 

We pre-train each model with self-supervised learning and fine-tune on FF++, we use the performances on FF++ HQ, FF++ LQ (same setting as \Tref{table:ff++}), and the average cross-datasets performances (same setting as \Tref{table:cross_dataset}) as metrics for evaluation. As the dimension of ResNet50 output is 2048, we double the width of its backend.  

Compared with 4-layers MSTCN backend, 4-layers transformer achieves about 1 percent AUC gain in FF++ LQ dataset and more than 2 percent AUC gain in cross dataset transfer. It verifies the advantage of the transformer. The ResNet50 based models outperform ResNet18, which indicates that a larger model will be helpful to capture more precise and detailed lip movements. When increasing the depth of the backend transformer from 6-layers to 8-layers, the performance of the ResNet18 model stops improving but the ResNet50 model still improves. This might indicate that the gain from deepening the backend is restricted by frontend scale and backend width.
Eventually, we choose ResNet18 (same setting as Lipforensics for a fair comparison) as frontend and 6 layers transformer as backend for the balance of efficiency and accuracy.

\begin{table*}[t]
\begin{center}
\setlength{\tabcolsep}{6mm}{
\begin{tabular}{l c c c c c}\hline
Frontend & backend & Total parameters & FF++ HQ & FF++ LQ & Avg Cross   \\ \hline
ResNet18 & L4-MSTCN & 36.0M & 98.8 & 95.4 & 84.5  \\
ResNet18 & L4-Transformer & 45.3M & 99.3 & 96.4 & 87.8  \\
ResNet18 & L6-Transformer & 62.1M & 99.6 & 96.7 & 88.4  \\
ResNet18 & L8-Transformer & 78.9M & 99.6 & 96.5 & 88.1  \\
ResNet50 & L4-MSTCN & 132.1M & 99.0 & 95.7 & 85.9  \\
ResNet50 & L4-Transformer & 162.1M & 99.5 & 96.6 & 87.4  \\
ResNet50 & L6-Transformer & 229.2M & 99.7 & 97.0 & 88.7  \\
ResNet50 & L8-Transformer & 296.3M & 99.7 & 97.1 & 88.9  \\
\hline 
\end{tabular}
}
\end{center}
\vspace{-1em}
\caption{Ablation study of network architecture combination of frontend and backend. The transformer backend outperforms MSTCN with the same layer number by a remarkable improvement. The model with larger frontend scale shows better performance in generalization. }
\label{table:ab_layers}
\vspace{-1em}
\end{table*}

\subsubsection{Effect of Pre-training Dataset Scale}

\begin{figure}[t]
   \centering
   \includegraphics[width=\linewidth]{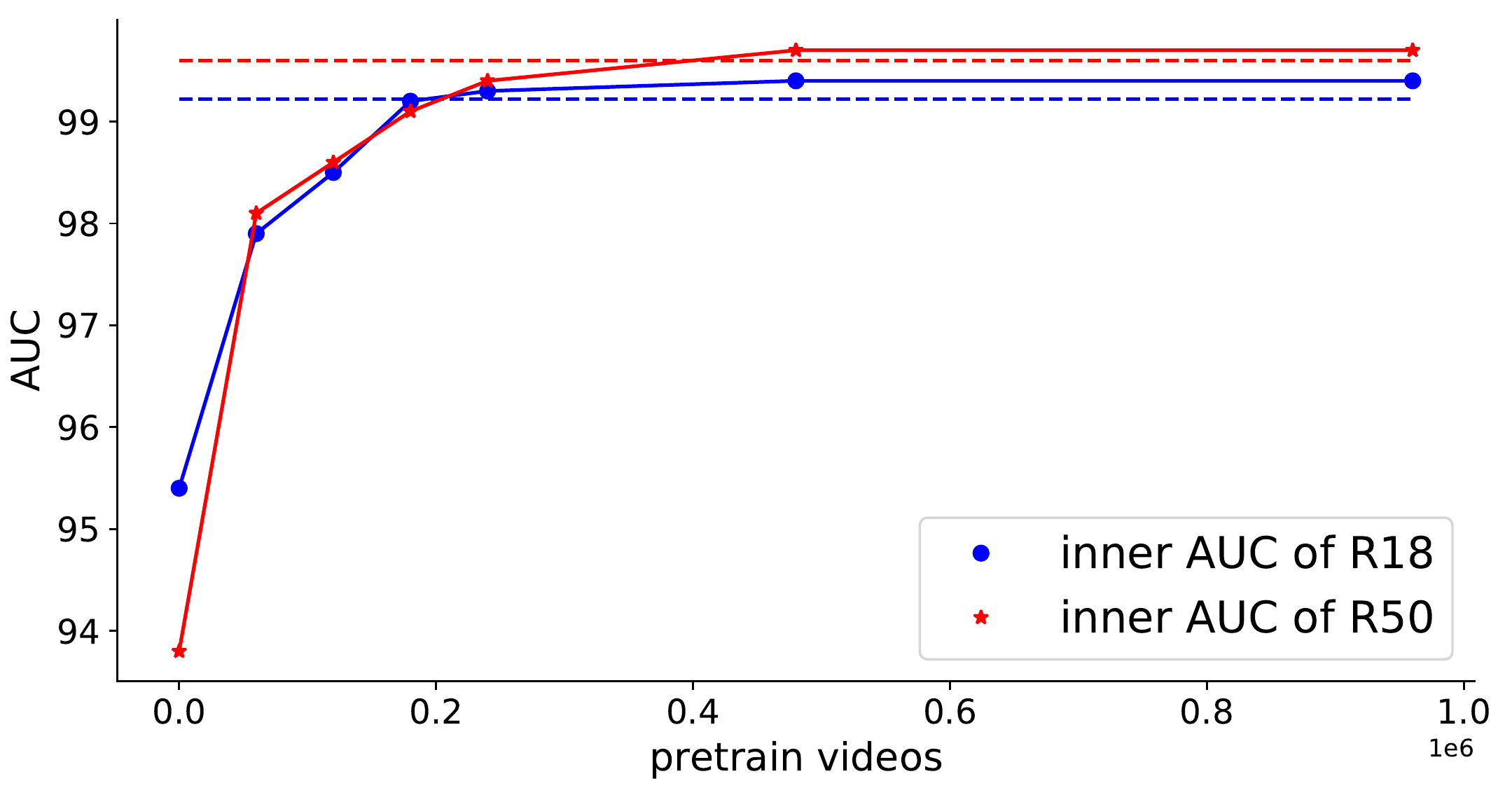}
   \caption{The relation between scale of pre-train data and test AUC on FF++ HQ dataset. }
   \label{figure:fig_data_c23}
 \vspace{-1em}
 \end{figure}
 
\begin{figure}[t]
   \centering
   \includegraphics[width=\linewidth]{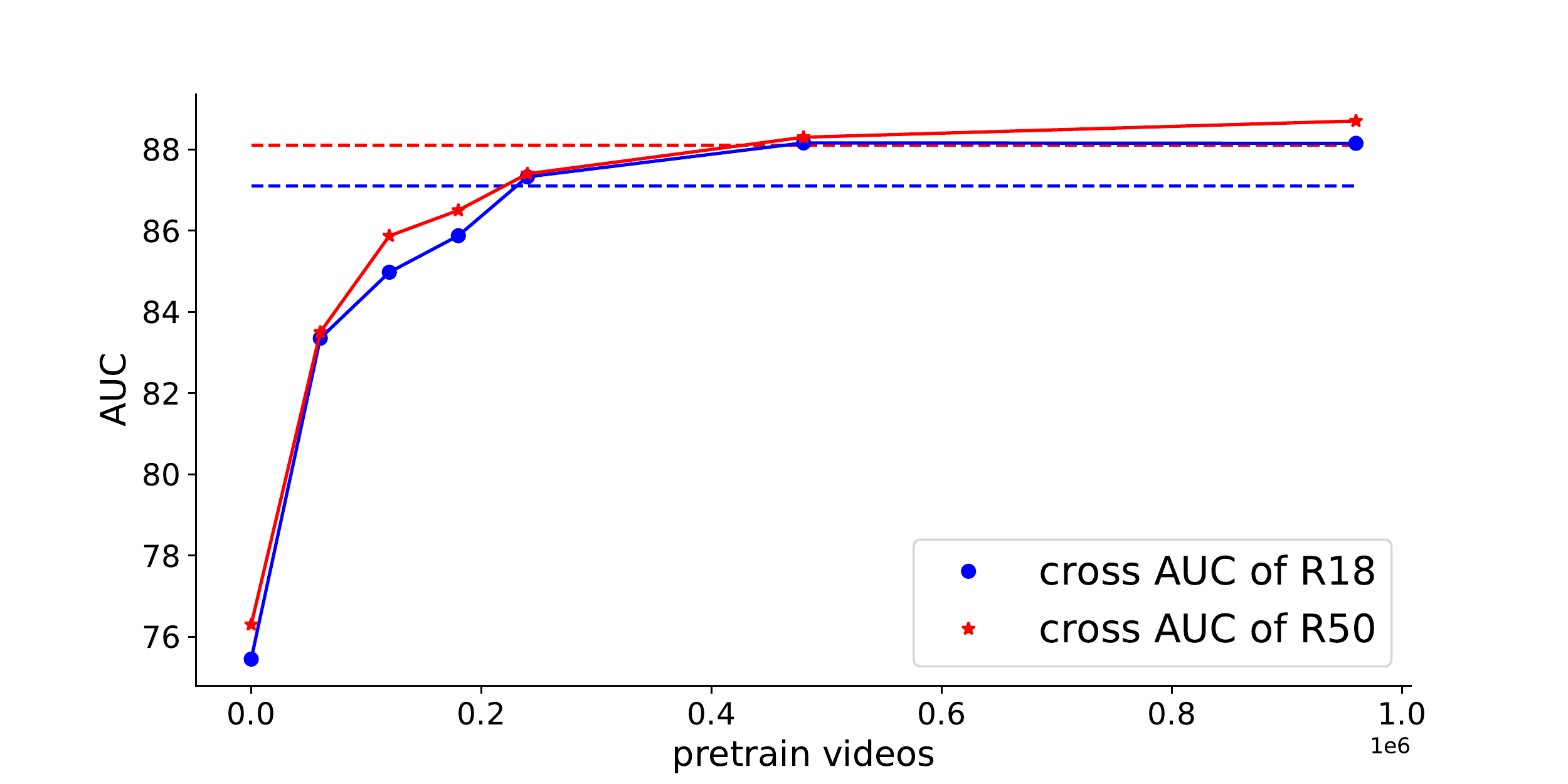}
   \caption{The relation between scale of pre-training data and average cross datasets AUC fine-tuned on FF++ dataset. }
   \label{figure:fig_data_cross}
    \vspace{-1em}
 \end{figure}

Compared to the supervised manner, self-supervised learning can leverage a larger amount of training data. To be specific, the labeled dataset LRW for lipreading task contains about 500,000 video clips of 500 words with precise boundary annotation. It only covers a limited part of real speech scenes but costs expensive human efforts in annotating. Contrastly, self-supervised learning only requires natural speech videos, which are very easy to collect. Empowered by self-supervised pre-train, we can improve the deepfake detection accuracy without requiring extra fake videos as training data. In this part, we give a simple investigation of how the quantity of unlabeled real video for pre-train affects the model's accuracy and the cross-dataset ability. We draw the relation curves in \Fref{figure:fig_data_c23} and \Fref{figure:fig_data_cross}. The blue lines are ResNet18 based models and the red lines are ResNet50 based models, we also plot the AUC of the model pre-trained on the LRW dataset with horizontal dotted lines. The abscissa of 0 represents that the model has not been pre-trained.

The curves in \Fref{figure:fig_data_c23} and \Fref{figure:fig_data_cross} validate that the models pre-trained with more original videos could improve both in-dataset and cross-datasets accuracies. When pre-trained with 480,000 video clips, the performance exceeds the same network pre-trained with 500,000 labeled video clips from LRW dataset. And the performance can still improve with the increase in training videos. From the comparison between the ResNet18 based model and the ResNet50 based model, we observe that without pre-train, the larger model does not outperform the smaller model on FF++ HQ while achieving only minor ascendency for generalization. But the models obtain obvious gains from pre-training. The AUC of the ResNet50 based model grows faster and saturates slower than ResNet18 due to the larger model scale. In conclusion, we demonstrate that larger-scale pre-training helps boost the models' performances. This highlights another advantage of our self-supervised method, that is, significantly reducing the data annotation costs to provide a larger scale of training data. However, larger-scale training data inevitably leads to higher computing resources demands, which might be a limitation.

\vspace{-0.5em} 
\subsubsection{Robustness of Pre-train Features}

\begin{figure*}[t]
   \centering
   \includegraphics[width=\linewidth]{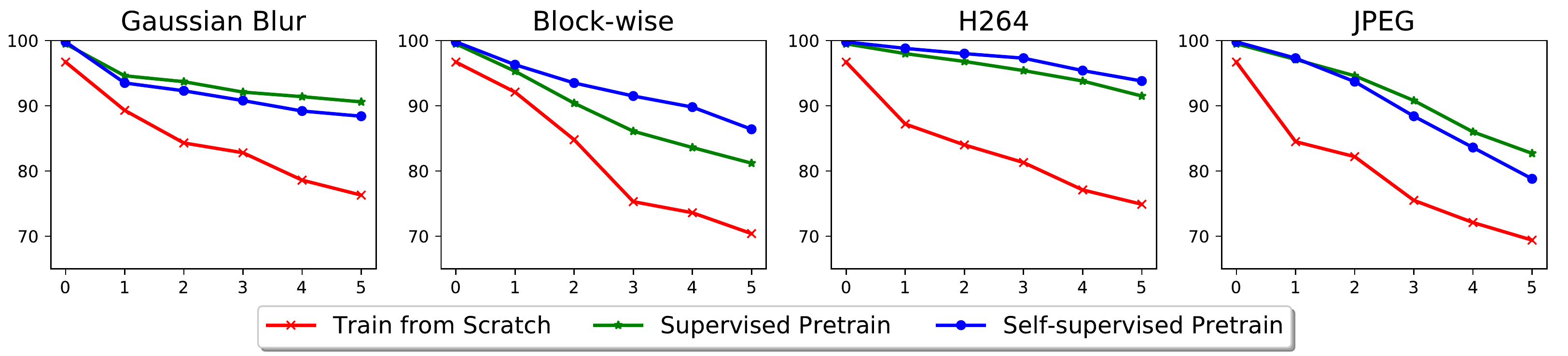}
   \caption{Video-level AUC scores of the spatial-temporal network as a function of the severity level for four types of perturbations: Gaussian Blur, Block-wise distortion, H264 Compression and Pixelation Distortion. We compare the robustness of models with each pre-train strategy.}
   \label{figure:fig_rob}
 \end{figure*}

 \begin{figure}[t]
   \centering
   \includegraphics[width=\linewidth]{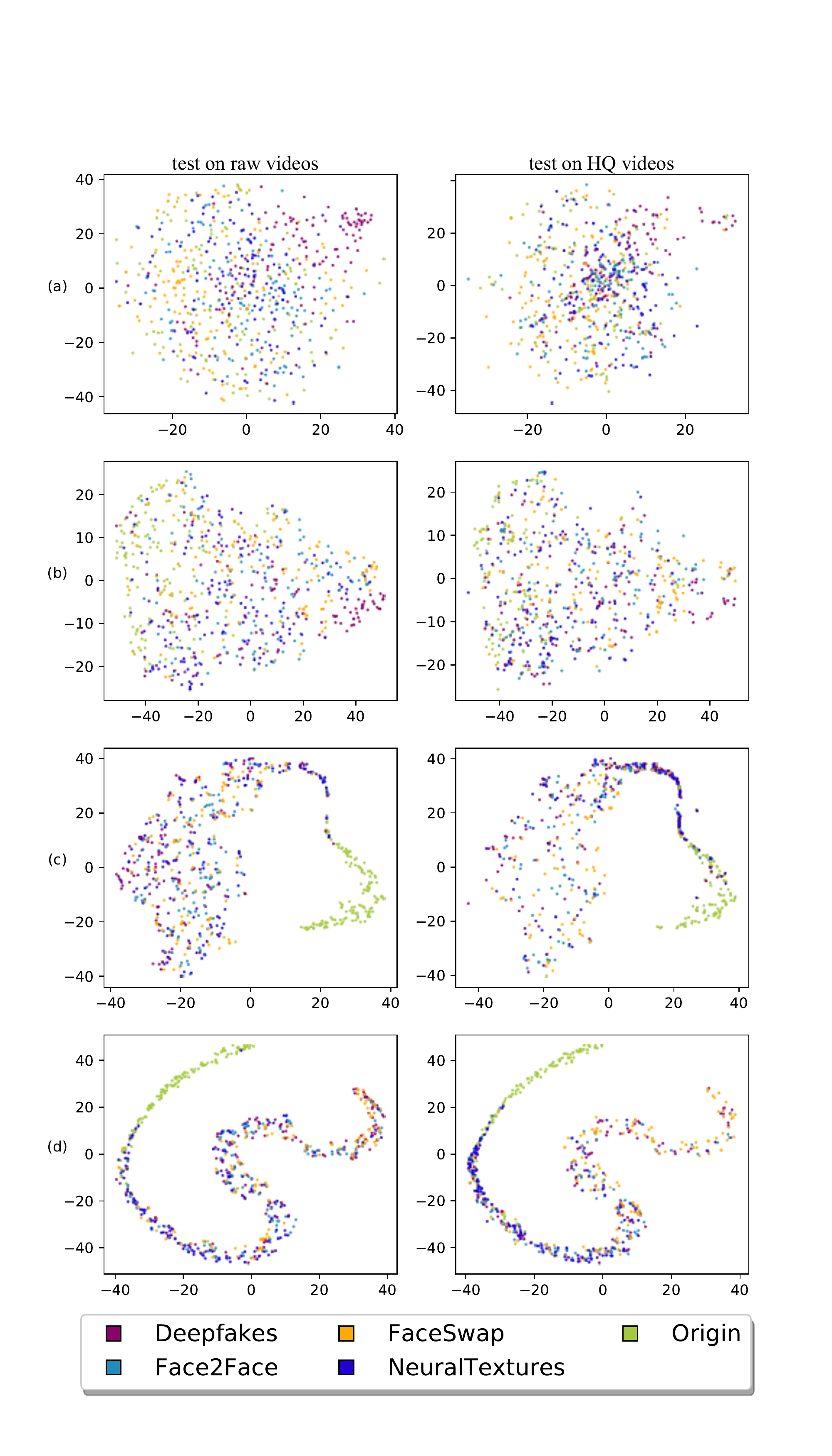}
   \caption{Visualization of features with t-SNE.  Features from self-supervise pre-trained model are less affected by compression. 
   }
   \label{figure:fig_aug}
 \end{figure} 
 
As aforementioned, the video in real-world scenarios may be accompanied by various disturbances, which requires that the detection model should be robust to various degradations of video quality. In our framework, we use massive real videos with various definitions to pre-train the video encoder. In this way, the learned representations of lip movement would be more robust to corruption. When fine-tuned for deepfake detection, we freeze the frontend of the video encoder so that the perturbations in videos would be reduced in the intermediate features. To evident the robustness of our model, we visualize the features of clean videos and corrupted videos extracted from the frontend and backend by our model. 

We visualize the features from the self-supervised pre-trained model and the train-from-scratch model in \Fref{figure:fig_aug}. The models are trained with FF++ Raw videos. The left column shows t-SNE embedded features from the Raw test set and the right column shows features from the HQ test set(compressed with H264 codec) transformed with Raw t-SNE embeddings. Row (a) and row (c) illustrate frontend and backend features of the non-pre-trained model separately, row (b) and row(d) are corresponding features of the self-supervised pre-trained model. We eventually observe that the distribution of the features trained with audio-visual contrastive learning changes more slightly compared to the train-from-scratch one. As a consequence, we believe that the deepfake detection models with self-supervised pretrain are less susceptible to interference from video compression.

Further, we investigate the robustness of our methods to various video corruptions. We take four common corruption types to simulate various degradations of definition: adding block-wise distortions, Gaussian blurring, JPEG compression, and H264 video compression, each corruption with five severity levels (details refer to \cite{Jiang2020DeeperForensics10AL}). In \Fref{figure:fig_rob}, we give the AUC curves under different corruption types. In subjects of Block-wise distortion and video compression, our self-supervised method performs better robustness than supervised pre-train, and in other two subjects, our method also achieves comparable performances.

\vspace{-0.5em} 
\section{Conclusion}
\vspace{-0.5em} 

In this paper, we propose a self-supervised pre-train method using audio-visual contrastive learning for improving the robustness and transferability of deepfake detection. Our method extracts a generic representation of local lip movement by 3D convolution and 2D ResNet and captures the long-term incoherence of lip movement with a transformer. Extensive experiments show that the generic pre-trained model is effective for deepfake detection. More importantly, with the self-supervised pre-trained framework, the accuracy and generalization ability can be possibly improved by scaling up model size and data. We believe the potential of self-supervised learning can be further explored for deepfake detection in the future.

{\small
\bibliographystyle{ieee_fullname}
\bibliography{egbib}
}

\end{document}